\documentclass[letterpaper, 10 pt, conference]{ieeeconf}  

\IEEEoverridecommandlockouts                              

\usepackage{graphicx} 
\usepackage{cite}
\usepackage{caption}
\usepackage{booktabs}
\usepackage{booktabs} 
\usepackage{tabularx}
\usepackage{makecell}
\usepackage{longtable}
\usepackage{multirow} 
\usepackage{array} 
\usepackage{dirtree}
\usepackage{placeins}
\usepackage{amsmath} 
\usepackage{fancyhdr}
\fancypagestyle{withfooter}{
  
  \fancyhead[L]{}
  \fancyhead[R]{}
  \fancyfoot[C]{\footnotesize Presented at the 2025 IEEE ICRA Workshop on Field Robotics}
}

\title{\LARGE \bf
AquaticVision: Benchmarking Visual SLAM in Underwater Environment with Events and Frames
}

\author{Yifan Peng$^{\ast}$, Yuze Hong$^{\ast}$, Ziyang Hong, Apple Pui-Yi Chui, Junfeng Wu  
\thanks{*Equal contribution }
\thanks{This work was supported in part by NSFC under Grant
62273288 and in part by the Shenzhen Science and Technology Program under Grant JCYJ20220818103000001.}
\thanks{$^{1}$Yifan Peng, Yuze Hong, Ziyang Hong, Junfeng Wu are with School of Data Science, The Chinese University of Hong Kong, Shenzhen, P. R. China,  \textbf{Corresponding author:} Ziyang Hong, Email: hongziyang@cuhk.edu.cn}
\thanks{$^{2}$Apple Pui-Yi Chui is with School of Life Sciences,  The Chinese University of Hong Kong, Hongkong}
}

\begin{document}

\maketitle
\thispagestyle{withfooter}
\pagestyle{withfooter}

\begin{abstract}

Many underwater applications, such as offshore asset inspections, rely on visual inspection and detailed 3D reconstruction. Recent advancements in underwater visual SLAM systems for aquatic environments have garnered significant attention in marine robotics research. However, existing underwater visual SLAM datasets often lack groundtruth trajectory data, making it difficult to objectively compare the performance of different SLAM algorithms based solely on qualitative results or COLMAP reconstruction. In this paper, we present a novel underwater dataset that includes ground truth trajectory data obtained using a motion capture system. Additionally, for the first time, we release visual data that includes both events and frames for benchmarking underwater visual positioning. By providing event camera data, we aim to facilitate the development of more robust and advanced underwater visual SLAM algorithms. The use of event cameras can help mitigate challenges posed by extremely low light or hazy underwater conditions. The webpage of our dataset is https://sites.google.com/view/aquaticvision-lias.

\end{abstract}

\section{INTRODUCTION}

With the deepening exploration of oceans, underwater robotics applications have gained significant prominence. Visual SLAM technology provides robots with reliable localization and environmental perception capabilities in GPS-denied underwater environments at relatively low cost, serving as a crucial foundation for intelligent autonomous underwater robotics. Unlike terrestrial environments, visual SLAM in underwater scenarios faces substantially more challenges, including unpredictable lighting conditions, variable water clarity, and abundant unstructured textures. Consequently, conventional visual cameras alone cannot adequately address underwater challenges, prompting researchers to explore novel sensor technologies to enhance the robustness and accuracy of underwater visual SLAM systems.

In recent years, event camera has attracted considerable academic attention as bio-inspired visual sensors due to their high dynamic range (HDR), low latency, and high temporal resolution. Unlike traditional frame-based cameras that capture entire scenes at fixed frame rates, the event camera asynchronously records intensity changes at each pixel, generating data only when brightness variations exceed predetermined thresholds. This operating mechanism renders event camera data more robust in underwater environments, opening new possibilities for underwater visual SLAM to some extent. The complementary strengths of event camera and traditional camera may offer a meaningful pathway for developing more robust underwater visual SLAM systems.

Most existing underwater datasets focus on deep learning tasks such as semantic segmentation and object detection, as in \cite{islam2020semantic} \cite{pedersen2019detection}. Additionally, \cite{8794272} provides a dataset for underwater color correction and depth estimation. SQUID \cite{berman2020underwater} is a dataset for 3D scene reconstruction. Currently, there are few datasets available for validating underwater visual SLAM, with the dataset released in \cite{rahman2018sonar} usable for VIO system validation, but without providing groundtruth. Recently, the FLSea dataset proposed in \cite{randall2023flsea} contains stereo images and IMU data collected in real underwater environments and provides groundtruth, making it suitable for validating both SLAM and depth estimation algorithms. However, as mentioned in \cite{randall2023flsea}, the pose groundtruth in FLSea is not captured using motion capture systems and may contain slight imperfections. 

To provide more reliable groundtruth for underwater visual SLAM algorithms and to facilitate further advancements in this field, \textbf{this paper introduces the first underwater dataset that simultaneously incorporates event camera data, traditional camera frames, and IMU measurements. Furthermore, we release groundtruth trajectories tracked by a motion capture system,} enabling researchers to validate the localization accuracy and performance of their algorithms in underwater environments. 

\begin{figure*}[htbp]
    \centering
    \includegraphics[width=\linewidth]{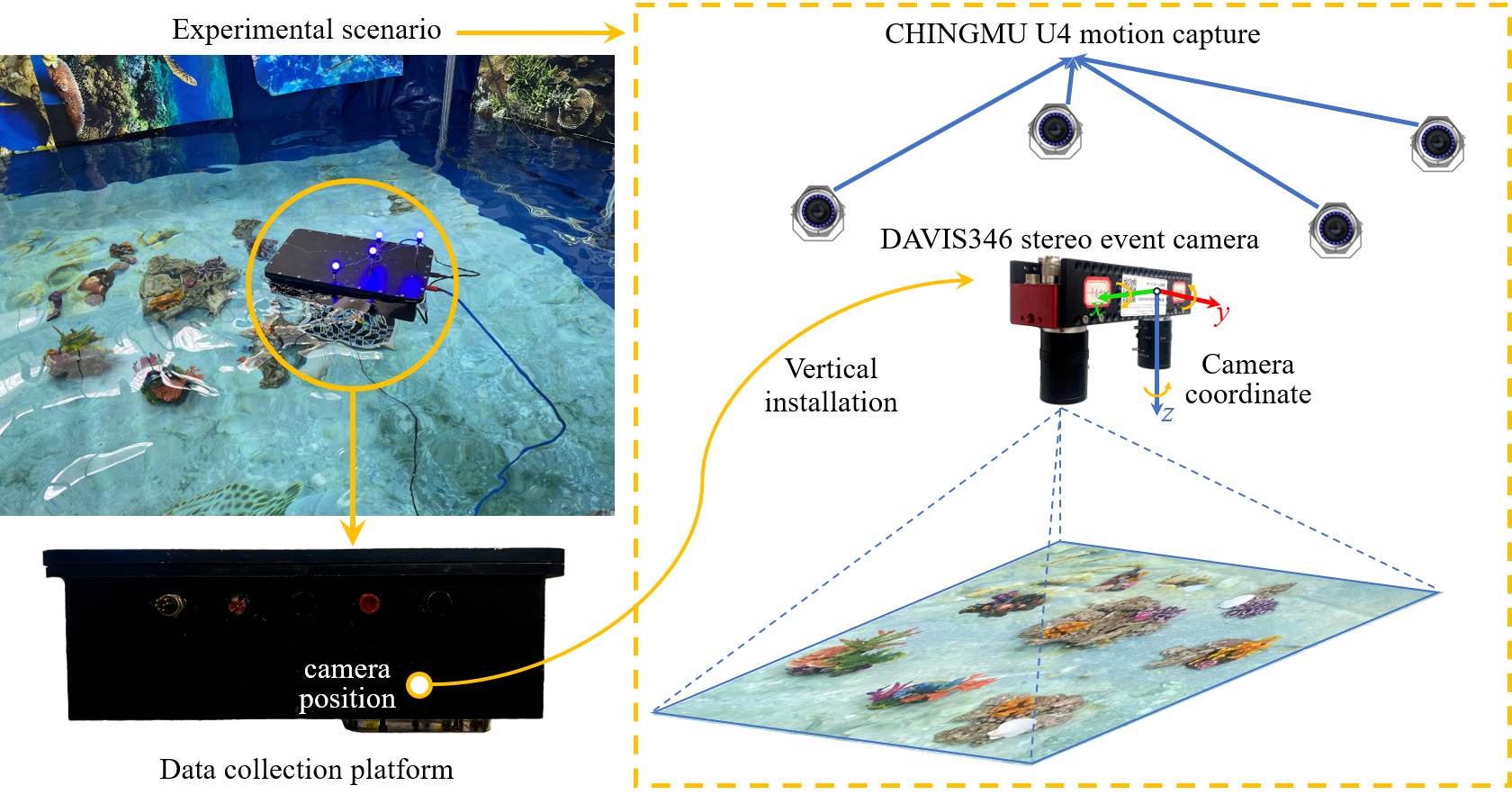}
    \captionsetup{justification=raggedright,singlelinecheck=false}
    \caption{Experimental scenario and underwater data collection platform. IMU coordinates align with camera one.}
    \label{fig1}
\end{figure*}


\section{Overview}

\subsection{Sensors setup}

The DAVIS346 Stereo Kit serves as our comprehensive data acquisition system. This sensor integrates dual DAVIS346-Mono event cameras capable of simultaneously capturing event streams and grayscale images. MPU-6500 IMU is embedded within the sensor, providing high-precision motion sensing capabilities with 3-axis gyroscope and 3-axis accelerometer measurements. 

The resulting dataset features high-temporal-resolution stereo event data sampled at 1000 Hz, complemented by stereo grayscale frames captured at 20 Hz with a spatial resolution of 346 × 260 pixels. Additionally, the system incorporates IMU data synchronized at 200 Hz. Timestamps between all sensors are synchronized in hardware. Besides, for all data sequences published in our dataset, the groundtruth trajectories are aquired by using CHINGMU-U4 motion capture system. It provides 6D pose measurements of a specific coordinates frame, which is defined by a set of marker attached on the top of data collection platform. The pose measurements are recorded at a rate of 120 Hz. Table \ref{tab1} summarizes all sensors used and their data publishing frequency.

\begin{table}[htbp]
  \centering
  \caption{Sensors used in the experiments}
  \label{tab1}
  \begin{tabularx}{0.48\textwidth}{>{\raggedright\arraybackslash}p{2.5cm}XXX}  
    \toprule  
    \textbf{Sensor} & \textbf{Type} & \textbf{Frequency} \\
    \midrule  
    Event camera& DAVIS346 & 1000Hz \\
    Frame camera & DAVIS346 & 20Hz/30Hz \\
    IMU & MPU-6500 & 200Hz/1000Hz \\
    Motion capture & CHINGMU-U4 & 120Hz \\
    \bottomrule  
  \end{tabularx}
\end{table}

\subsection{Experimental Scenario}

The experimental scene is a pool with dimensions of 4 × 3 × 2 meters, with a water depth of approximately 1 meter. The pool bottom is covered with a real seabed beach image as background, and multiple coral models are placed to simulate the underwater environment to the greatest extent possible. Vertically installed in the data collection platform with its z-axis oriented toward the pool bottom, the DAVIS346 event camera captures data as the platform navigates within a 3 × 2 meter planar area. Four motion capture cameras are set up above the pool to obtain groundtruth. Fig. \ref{fig1} illustrates the experimental scenario and the data collection platform. It should be noted that the motion capture equipments are not installed underwater, which is done to expand the robot's range of motion. The collection platform is actually floating on the water surface, with markers installed above it, higher than the water level.

Based on the visual feature characteristics of the scenes, we divide the collected 9 data sequences into two categories: easy and hard. Sequences 01 through 05 belong to the easy category, featuring clear visual characteristics that simulate underwater environments with good lighting and water quality. Sequences 06 through 09 belong to the hard category, including three unique underwater environmental challenges: darkness, HDR (High Dynamic Range), and blur. TABLE \ref{tab2} shows the characteristics of all nine data sequences. The with\_board or no\_board suffix indicates whether a calibration board is present underwater.

\begin{table*}[!ht]
  \centering
  \caption{Data sequences characteristics}
  \label{tab2}
  \begin{tabular*}{\textwidth}{@{\extracolsep{\fill}}ccccc@{}}
    \toprule
    \textbf{Sequence} & \textbf{Name} & \textbf{Category} & \textbf{Duration} & \textbf{Description} \\
    \midrule
     01 & Scan\_with\_board   & Easy & 93.32s & Clear water, corals evenly distributed, trajectory covers underwater area\\
     02 & Cross1\_with\_board & Easy & 109.00s & Clear water, corals centrally distributed, cross-pattern trajectory\\     
     03 & Cross2\_no\_board   & Easy & 68.03s & Clear water, corals evenly distributed, cross-pattern trajectory\\
     04 & Loop1\_with\_board  & Easy & 75.70s & Clear water, corals evenly distributed, loop-shaped trajectory \\
     05 & Loop2\_no\_board    & Easy & 59.92s & Clear water, corals centrally distributed, loop-shaped trajectory \\
     06 & Dark1\_with\_board  & Hard & 75.04s & Clear water, low-light condition\\
     07 & Dark2\_with\_board  & Hard & 74.30s & Clear water, low-light condition\\
     08 & HDR                 & Hard & 97.00s & High dynamic range lighting conditions creating challenging exposure variations \\
     09 & Blur                & Hard & 133.47s & Water turbidity affecting image clarity and feature detection \\
    \bottomrule
  \end{tabular*}
\end{table*}

\subsection{Calibration}

The DAVIS integrates both conventional imaging and event sensing capabilities on a unified pixel array, allowing for a streamlined calibration workflow. This architectural advantage enables researchers to employ established calibration frameworks like Kalibr \cite{furgale2013unified} on the standard image output, with the resulting calibration parameters being directly transferable to the event component without requiring separate calibration procedures. The extrinsic parameters between the camera and IMU were calibrated by using Kalibr \cite{furgale2013unified}.

\begin{table*}[!t]
  \centering
  \caption{ATE OF SLAM SYSTEMS ON SAMPLE SEQUENCES [m]}
  \label{tab3}
  \renewcommand{\arraystretch}{1.2}
  \setlength{\tabcolsep}{0.5em}
  \begin{tabular*}{\textwidth}{@{\extracolsep{\fill}}c|ccccccccc@{}}
    \toprule  
    \textbf{Method}  & \textbf{S01} & \textbf{S02} & \textbf{S03} & \textbf{S04} & \textbf{S05} & \textbf{S06} & \textbf{S07} & \textbf{S08} & \textbf{S09}  \\
    \midrule  
    VINS-Stereo & 0.0573 & 0.0616 & 0.0428 & 0.0489 & 0.0907 & 1.1629 & 0.0698 & failed & 0.0507 \\
    ORB-SLAM2   & 0.4231 & 0.1134 & 0.0864 & 0.4267 & 0.1712 & failed & failed & 0.1763 & 0.1546 \\
    ESVO2       & failed & failed & failed & failed & failed & failed & failed & failed & failed \\
    \bottomrule  
  \end{tabular*}
\end{table*}

\subsection{Data Format}

The proposed dataset is orgnized as follow:

\dirtree{%
.1 dataset\_name.
.2 calibration.
.3 cam0\_pinhole.yaml.
.3 cam1\_pinhole.yaml.
.3 davis\_imucam\_underwater.yaml.
.2 data\_sequences.
.3 01\_Scan\_with\_board.
.3 \ldots.
.3 09\_Blur.
}

In the calibration folder, there are three calibration files in YAML format, which contain the intrinsic parameters of the left camera (cam0) and right camera (cam1), as well as the extrinsic parameters between the cameras and the IMU. The data\_sequences folder contains data for 9 sequences.

Each sequence provides sensor data, groundtruth, and grayscale images. Event, image, and IMU data are provided in the rosbag format, including two types of rosbags: one with images at 30Hz and IMU at 1000Hz, and another with images at 20Hz and IMU at 200Hz. Users can choose the appropriate rosbag for their algorithms and perform downsampling operations when needed. The baseline trajectory is provided in the CSV file. Additionally, we have released a groundtruth file in TUM format to facilitate evaluation based on the evo tool. Furthermore, the left and right camera grayscale images at 30Hz frame rate have been extracted and stored separately in the l1 and r1 folders, accompanied by a timestamp alignment file. The data sequence is orgnized as follow:

\dirtree{%
.1 id\_sequence\_name.
.2 data rosbag.
.3 name\_imu\_1000hz\_images\_30hz.bag.
.3 name\_imu\_200hz\_images\_20hz.bag.
.2 groundtruth.
.3 gt.csv.
.3 gt.tum.
.2 Stereo images.
.3 l1.
.3 r1.
.3 timestamp\_pairs.txt.
}

\subsection{Baseline trajectory using visual SLAM}

Various mainstream SLAM systems are tested across nine sequences from the proposed dataset, with detailed characteristics presented in Table II. To assess performance, we employ the Absolute Trajectory Error (ATE) as our primary metric. For consistent comparison, we utilize the evo tool to align each estimated trajectory with its corresponding ground truth, thereby calculating precise ATE measurements. The quantitative results are shown in Table \ref{tab3}.

To evaluate the performance of frame-based visual SLAM in underwater environments, Vins-Stereo-fusion \cite{qin2019a} and ORB-SLAM2 \cite{mur2017orb} are used for testing. The testing of Vins is conducted using rosbags with IMU data at 200Hz and images at 20Hz, while the testing of ORB-SLAM2 utilizes grayscale images at 30Hz. It should be specifically noted that the purpose of selecting ORB-SLAM2 for testing is to assess the performance difference between visual odometry (VO) and visual-inertial odometry (VIO) on the proposed underwater dataset. The state-of-the-art event-based SLAM algorithm, ESVO2 \cite{10912788}, is also tested to evaluate the performance of event camera in challenging underwater environments. 

\begin{figure}[htbp]
    \centering
    \includegraphics[width=\linewidth]{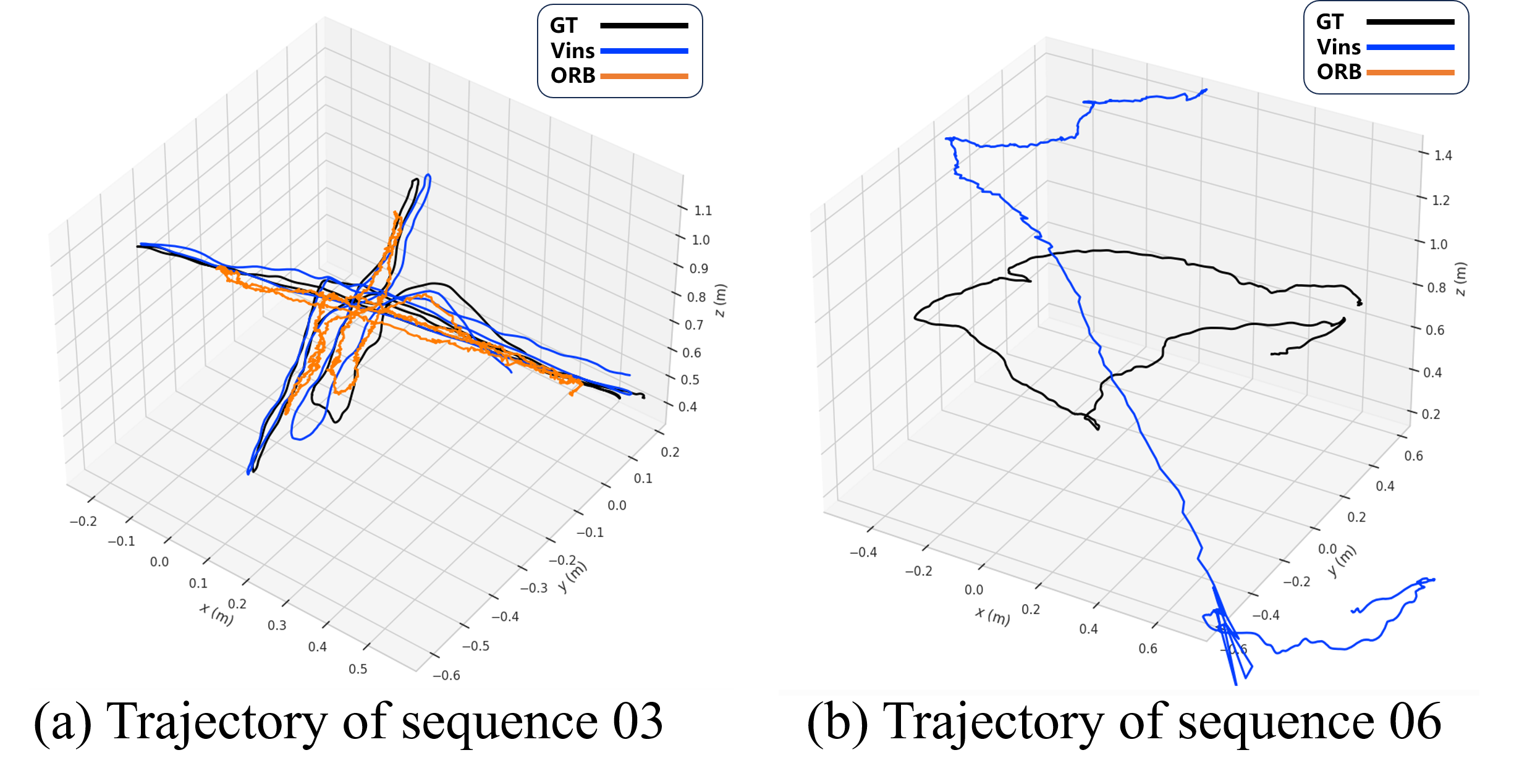}
    \captionsetup{justification=raggedright,singlelinecheck=false}
    \caption{Estimated and groundtruth (GT) trajectories of 2 sample sequences.}
    \label{fig2}
\end{figure}

\begin{figure*}[!t]
    \centering
    \includegraphics[width=\linewidth]{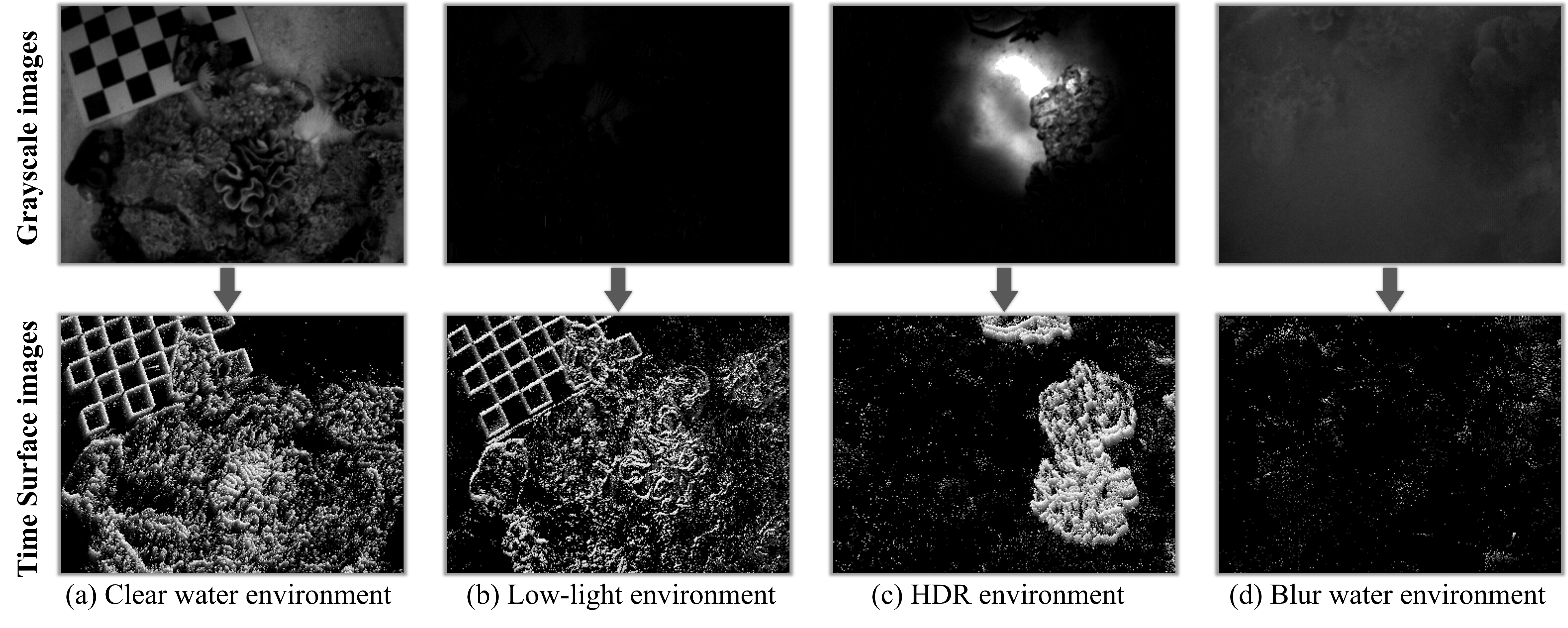}
    \captionsetup{justification=raggedright,singlelinecheck=false}
    \caption{Comparison of grayscale images and their corresponding Time Surface images under different underwater environmental conditions.}
    \label{fig3}
\end{figure*}

\noindent\textbf{\textit{1) Frame-based visual SLAM: }}

As shown in Table \ref{tab3}, test results based on the easy class sequences (S01-S05) indicate that the pose estimation accuracy of the VIO system (VINS-Stereo) is significantly higher than that of the VO system (ORB-SLAM2). This is because the presence of water waves and buoyancy causes high-frequency rapid jitter of the data collection platform during turns. On one hand, this violates the constant velocity motion model assumption relied upon by ORB-SLAM2, making pose estimation no longer accurate. On the other hand, under the premise of similar features in underwater environment and lower camera resolution, small-range rapid movements can cause degradation in visual pose estimation. In fact, based on what was mentioned in SVin2 \cite{8967703}, this paper also applied CLAHE histogram equalization \cite{pizer1987adaptive} to enhance visual features in the input images, but still did not achieve better results. VINS-Stereo, with the assistance of IMU data, can effectively compensate for the deficiencies in visual information. The IMU can provide high-frequency attitude and acceleration measurements, maintaining accurate pose estimation in the short term even when visual features are scarce. For data sequences with smoother motion and without rapid turns (S03), both ORB-SLAM2 and VINS can run relatively well, as shown in Fig. \ref{fig2}(a).

In experiments based on hard class sequences, both Vins-Stereo and ORB-SLAM2 exhibited different degrees of failure. Under low-light conditions (S06-S07), ORB-SLAM2 struggled to extract effective ORB feature points, making it impossible to estimate trajectories. VINS-Stereo, on the other hand, uses a combination of Shi-Tomasi corner detection with optical flow tracking for matching, which can detect more reliable corners in low-contrast images, as shown in Fig. \ref{fig2}(b). Additionally, optical flow tracking utilizes inter-frame continuity rather than re-detecting and matching features, reducing dependency on single-frame image quality, thus enabling more continuous trajectories. However, in HDR environment (S08), only a small area is illuminated, causing the scene to be in a state of constant abrupt changes, which can cause optical flow tracking to fail and affect VINS's performance. For blurry environment (S09), although ORB-SLAM2 had positioning errors of around 0.15m, it could not operate stably throughout the entire sequence, experiencing tracking loss during blurry frames. It was only through its effective relocalization capability that it could obtain pose estimation results for clear frames.

To sum up, IMU data is essential for the application of visual SLAM system in underwater scenarios. Meanwhile, to ensure the performance of underwater visual odometry, it is also important to extract more robust visual feature points.

\noindent \textbf{\textit{2) Event-based visual SLAM:}}

As shown in Table \ref{tab3}, ESVO2 failed on all data sequences. The main reasons for ESVO2's failure can be attributed to: ESVO2 requires extremely high calibration precision, especially when distortion correction is not accurate enough, which significantly affects the system's precision. Although the Kalibr calibration tool used in this paper can effectively obtain intrinsic and extrinsic parameters on land, it does not include specific optimizations for underwater calibration. Underwater, light passes through three layers of refraction and reflection - water, transparent panel, and air - which to some extent affects the accuracy of camera calibration. Additionally, due to the low resolution of the DAVIS346 camera and our short stereo camera baseline, there were negative impacts on depth estimation results and overall accuracy. Finally, ESVO2 demands robust initialization with high-quality reconstruction, however, the cluttered underwater environment textures often lead to reconstruction failures, affecting system operation. It should be noted that this does not mean ESVO2 lacks good pose estimation capability, but only indicates that it cannot effectively address challenges in underwater scenes under this paper's experimental setup. In fact, ESVO2 has demonstrated excellent performance on land when using the long-baseline, high-resolution solution.

\subsection{Event Imaging Under Adverse Conditions}

Event cameras have unique advantages in visual representation of complex underwater scenes. Time Surface (TS) is a commonly used event representation method, with the following principle:

\begin{equation}
\mathcal{T}(\mathbf{x}, t) \doteq \exp\left(-\frac{t - t_{\text{last}}(\mathbf{x})}{\eta}\right),
\end{equation}

where $t_{\text{last}}$ is the timestamp of the last event at each pixel coordinate $\mathbf{x} = (u, v)^\top$, and $\eta$ is a parameter controlling the decay rate. Time Surface uses an exponential decay kernel to emphasize recent events while attenuating past events.

\begin{figure}[htbp]
    \centering
    \includegraphics[width=\linewidth]{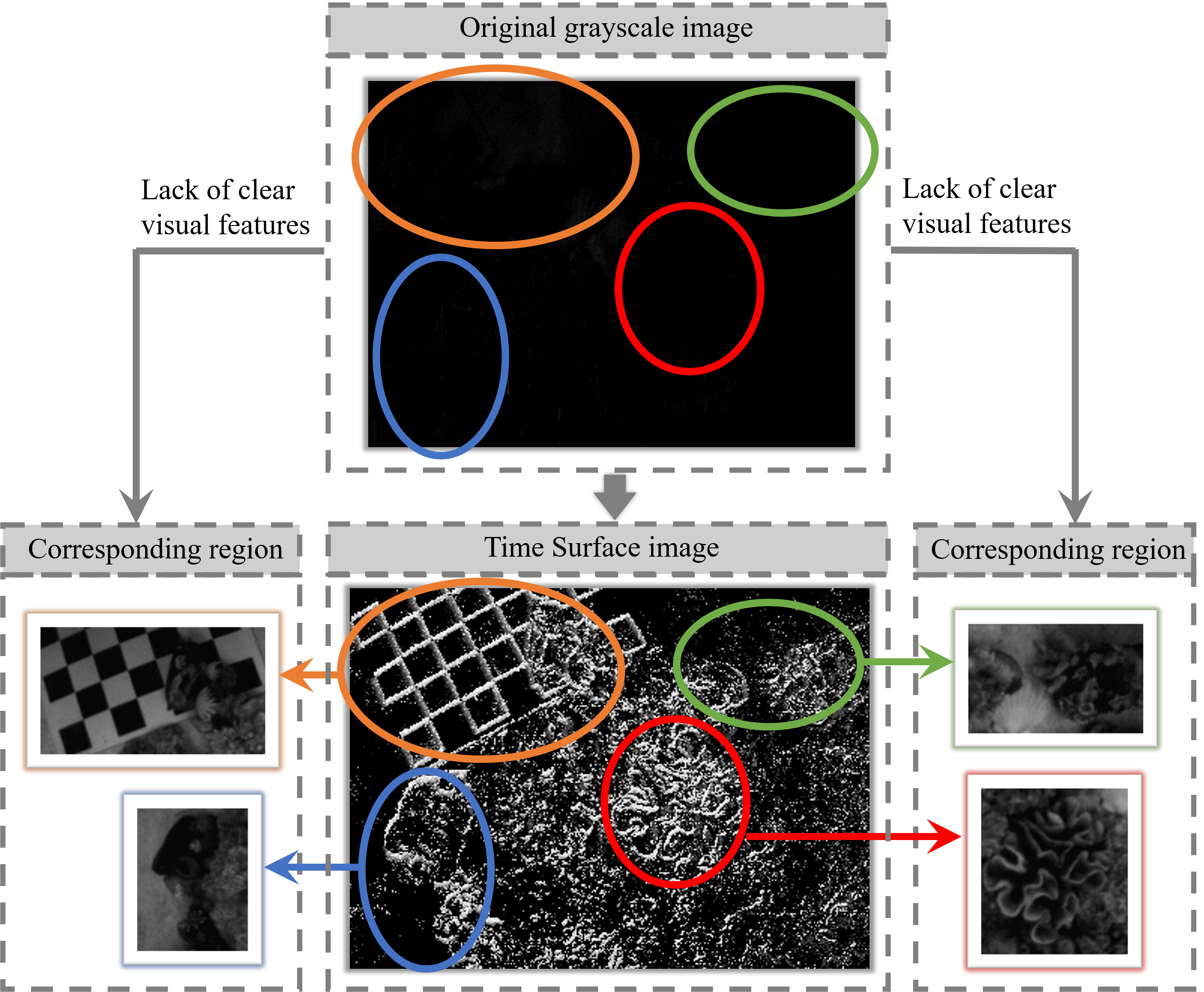}
    \captionsetup{justification=raggedright,singlelinecheck=false}
    \caption{Details comparison of Time Surface image and original grayscale imaging features in low-light underwater environment (Sequence 06).}
    \label{fig4}
\end{figure} 

The dataset published in this paper includes four scenarios: clear water, low light, HDR, and turbid water. Fig. \ref{fig3} (a) shows the clear water condition, where the corresponding Time Surface effectively reflects the environmental texture features. For coral regions with low contrast, Time Surface can also represent their surface textures and contours. A detailed comparison of Time Surface images and original grayscale imaging features in low-light underwater environments is shown in Fig. \ref{fig4}. Fig. \ref{fig3} (b) and (c) represent challenging lighting conditions. Evidently, Time Surfaces generated from event data demonstrate strong robustness to lighting conditions, effectively presenting environmental textures in low-light or HDR scenes.

\noindent \textbf{Failure case:} As shown in Fig. \ref{fig3}(d), Time Surfaces fail when water becomes turbid. This is because numerous impurities obscure environmental details, preventing event cameras from properly capturing changes in illumination. This phenomenon indicates that event cameras do not necessarily perform better than conventional cameras in certain scenarios.

\noindent \textbf{Going beyond TS image: } While Time Surface imaging effectively reveals features in clear and low-light conditions, more robust event representations are needed for challenging underwater environments. To be more specific, Motion-compensated Event Frame method \cite{gallego2017accurate}\cite{gallego2018unifying}  that warps event data to a reference frame using motion models can significantly reduce the influence of suspended particles by aligning events from real objects while dispersing random noise. Event Packet representations \cite{9138762} preserves temporal precision by aggregating events within time windows, enabling batch operations that better distinguish between consistent features and sporadic noise in turbid waters. Voxel Grid \cite{huang2023event} approach encodes events into 3D space-time tensor representations, allowing for structured spatial and temporal integration that enhances signal-to-noise ratio in environments with varying visibility. By combining the advantages of these methods, adaptive systems can be developed for different underwater conditions, enabling more reliable perception and localization in challenging underwater environments.

\section{CONCLUSION}

This paper presents an underwater dataset containing event, frame, and IMU data, along with groundtruth trajectories for evaluation purpose. The dataset is categorized into "easy" and "hard" classes, where sequences in the easy class were collected under good water quality and lighting conditions, while sequences in the hard class include special underwater conditions such as low light, high dynamic range (HDR), and turbidity. Furthermore, based on this dataset, we evaluated mainstream frame-based visual SLAM algorithms and  a state-of-the-art event-based SLAM algorithm in underwater scenarios. The analysis of advantages and limitations of event camera in underwater environments points to potential future directions for underwater SLAM research. In the future, we will continue to improve and update this dataset, providing more realistic underwater scenes and more accurate calibration data, striving to contribute a comprehensive benchmark for underwater robotics research.

\bibliographystyle{ieeetr}
\bibliography{paper}

\begin{thebibliography}{10}

\bibitem{islam2020semantic}
M.~J. Islam, C.~Edge, Y.~Xiao, P.~Luo, M.~Mehtaz, C.~Morse, S.~S. Enan, and J.~Sattar, ``Semantic segmentation of underwater imagery: Dataset and benchmark,'' in {\em 2020 IEEE/RSJ International Conference on Intelligent Robots and Systems (IROS)}, pp.~1769--1776, IEEE, 2020.

\bibitem{pedersen2019detection}
M.~Pedersen, J.~Bruslund~Haurum, R.~Gade, and T.~B. Moeslund, ``Detection of marine animals in a new underwater dataset with varying visibility,'' in {\em Proceedings of the IEEE/CVF conference on computer vision and pattern recognition workshops}, pp.~18--26, 2019.

\bibitem{8794272}
K.~A. Skinner, J.~Zhang, E.~A. Olson, and M.~Johnson-Roberson, ``Uwstereonet: Unsupervised learning for depth estimation and color correction of underwater stereo imagery,'' in {\em 2019 International Conference on Robotics and Automation (ICRA)}, pp.~7947--7954, 2019.

\bibitem{berman2020underwater}
D.~Berman, D.~Levy, S.~Avidan, and T.~Treibitz, ``Underwater single image color restoration using haze-lines and a new quantitative dataset,'' {\em IEEE transactions on pattern analysis and machine intelligence}, vol.~43, no.~8, pp.~2822--2837, 2020.

\bibitem{rahman2018sonar}
S.~Rahman, A.~Q. Li, and I.~Rekleitis, ``Sonar visual inertial slam of underwater structures,'' in {\em 2018 IEEE International Conference on Robotics and Automation (ICRA)}, pp.~5190--5196, IEEE, 2018.

\bibitem{randall2023flsea}
Y.~Randall and T.~Treibitz, ``Flsea: Underwater visual-inertial and stereo-vision forward-looking datasets,'' {\em arXiv preprint arXiv:2302.12772}, 2023.

\bibitem{furgale2013unified}
P.~Furgale, J.~Rehder, and R.~Siegwart, ``Unified temporal and spatial calibration for multi-sensor systems,'' in {\em 2013 IEEE/RSJ International Conference on Intelligent Robots and Systems}, pp.~1280--1286, IEEE, 2013.

\bibitem{qin2019a}
T.~Qin, J.~Pan, S.~Cao, and S.~Shen, ``A general optimization-based framework for local odometry estimation with multiple sensors,'' 2019.

\bibitem{mur2017orb}
R.~Mur-Artal and J.~D. Tard{\'o}s, ``Orb-slam2: An open-source slam system for monocular, stereo, and rgb-d cameras,'' {\em IEEE transactions on robotics}, vol.~33, no.~5, pp.~1255--1262, 2017.

\bibitem{10912788}
J.~Niu, S.~Zhong, X.~Lu, S.~Shen, G.~Gallego, and Y.~Zhou, ``Esvo2: Direct visual-inertial odometry with stereo event cameras,'' {\em IEEE Transactions on Robotics}, vol.~41, pp.~2164--2183, 2025.

\bibitem{8967703}
S.~Rahman, A.~Q. Li, and I.~Rekleitis, ``Svin2: An underwater slam system using sonar, visual, inertial, and depth sensor,'' in {\em 2019 IEEE/RSJ International Conference on Intelligent Robots and Systems (IROS)}, pp.~1861--1868, 2019.

\bibitem{pizer1987adaptive}
S.~M. Pizer, E.~P. Amburn, J.~D. Austin, R.~Cromartie, A.~Geselowitz, T.~Greer, B.~ter Haar~Romeny, J.~B. Zimmerman, and K.~Zuiderveld, ``Adaptive histogram equalization and its variations,'' {\em Computer vision, graphics, and image processing}, vol.~39, no.~3, pp.~355--368, 1987.

\bibitem{gallego2017accurate}
G.~Gallego and D.~Scaramuzza, ``Accurate angular velocity estimation with an event camera,'' {\em IEEE Robotics and Automation Letters}, vol.~2, no.~2, pp.~632--639, 2017.

\bibitem{gallego2018unifying}
G.~Gallego, H.~Rebecq, and D.~Scaramuzza, ``A unifying contrast maximization framework for event cameras, with applications to motion, depth, and optical flow estimation,'' in {\em Proceedings of the IEEE conference on computer vision and pattern recognition}, pp.~3867--3876, 2018.

\bibitem{9138762}
G.~Gallego, T.~Delbrück, G.~Orchard, C.~Bartolozzi, B.~Taba, A.~Censi, S.~Leutenegger, A.~J. Davison, J.~Conradt, K.~Daniilidis, and D.~Scaramuzza, ``Event-based vision: A survey,'' {\em IEEE Transactions on Pattern Analysis and Machine Intelligence}, vol.~44, no.~1, pp.~154--180, 2022.

\bibitem{huang2023event}
K.~Huang, S.~Zhang, J.~Zhang, and D.~Tao, ``Event-based simultaneous localization and mapping: A comprehensive survey,'' {\em arXiv preprint arXiv:2304.09793}, 2023.

\end{thebibliography}

\end{document}